\documentclass[letterpaper, 10 pt, conference]{ieeeconf}  
\usepackage{graphicx}
\usepackage{color}
\usepackage{multirow}
\usepackage{hyperref}
\usepackage{mathtools}
\usepackage{float}
\usepackage{bm}
\usepackage{amsmath,amssymb}
\usepackage[font=small,skip=3pt]{caption}
\input{macro_marchand_survey.sty}
\usepackage[nomessages]{fp}
\usepackage[all=normal,paragraphs=normal,floats=normal,mathspacing=normal,
wordspacing=tight,charwidths=tight,mathdisplays=tight,
leading=normal]{savetrees}
\graphicspath{{images/}}

\DeclareMathOperator*{\argmin}{arg\,min}

\IEEEoverridecommandlockouts                              

\newcommand{\modif}[1]{{\textcolor{black}{#1}}}
\title{\LARGE \bf
L6DNet: Light 6 DoF Network for Robust and\\ Precise Object Pose Estimation with Small Datasets$^*$}

\author{Mathieu Gonzalez$^{1}$, Amine Kacete$^{1}$, Albert Murienne$^{1}$ and Eric Marchand$^{2}$
\thanks{$^{*}$This work benefited from State aid managed by the National
Research Agency under the future investment program bearing the reference ANR-17-
RHUS-0005 (FollowKnee project).}
\thanks{$^{1}$ Mathieu Gonzalez, Amine Kacete and Albert Murienne are with the Institute of Research and Technology b$<>$com, Rennes, France,
        {\tt\small \{mathieu.gonzalez,amine.kacete, albert.murienne\}@b-com.com}}%
\thanks{$^{2}$ Eric Marchand is with Univ Rennes, Inria, IRISA, CNRS, Rennes, France,
        {\tt\small Eric.Marchand@irisa.fr}}%
}

\usepackage{hyperref}
\begin{document}
\onecolumn
\begin{center}
This paper has been accepted for publication in \textit{IEEE Robotics and Automation Letters}.
~\\
~\\
~\\
DOI: \href{https://dx.doi.org/10.1109/LRA.2021.3062605}{10.1109/LRA.2021.3062605} 
~\\
IEEE Xplore: \href{https://ieeexplore.ieee.org/document/9364353}{https://ieeexplore.ieee.org/document/9364353}
~\\
~\\
~\\
~\\
© 2021 IEEE.  Personal use of this material is permitted.  Permission from IEEE must be obtained for all other uses, in any current or future media, including reprinting/republishing this material for advertising or promotional purposes, creating new collective works, for resale or redistribution to servers or lists, or reuse of any copyrighted component of this work in other works.
\end{center}
\twocolumn
\maketitle
\thispagestyle{empty}
\pagestyle{empty}

\begin{abstract}

Estimating the 3D pose of an object is a challenging task that can be considered within augmented reality or robotic applications. In this paper, we propose a novel approach to perform 6 DoF object pose estimation from a single RGB-D image. We adopt a hybrid pipeline in two stages: data-driven and geometric respectively. The data-driven step consists of a classification CNN to estimate the object 2D location in the image from local patches, followed by a regression CNN trained to predict the 3D location of a set of keypoints in the camera coordinate system. To extract the pose information, the geometric step consists in aligning the 3D points in the camera coordinate system with the corresponding 3D points in world coordinate system by minimizing a registration error, thus computing the pose. Our experiments on the standard dataset LineMod show that our approach is more robust and accurate than state-of-the-art methods. The approach is also validated to achieve a 6 DoF positioning task by visual servoing.

\end{abstract}
~\\

\section{Introduction}
\begin{figure}[H]
    \centering
    \includegraphics[width=1.0\columnwidth]{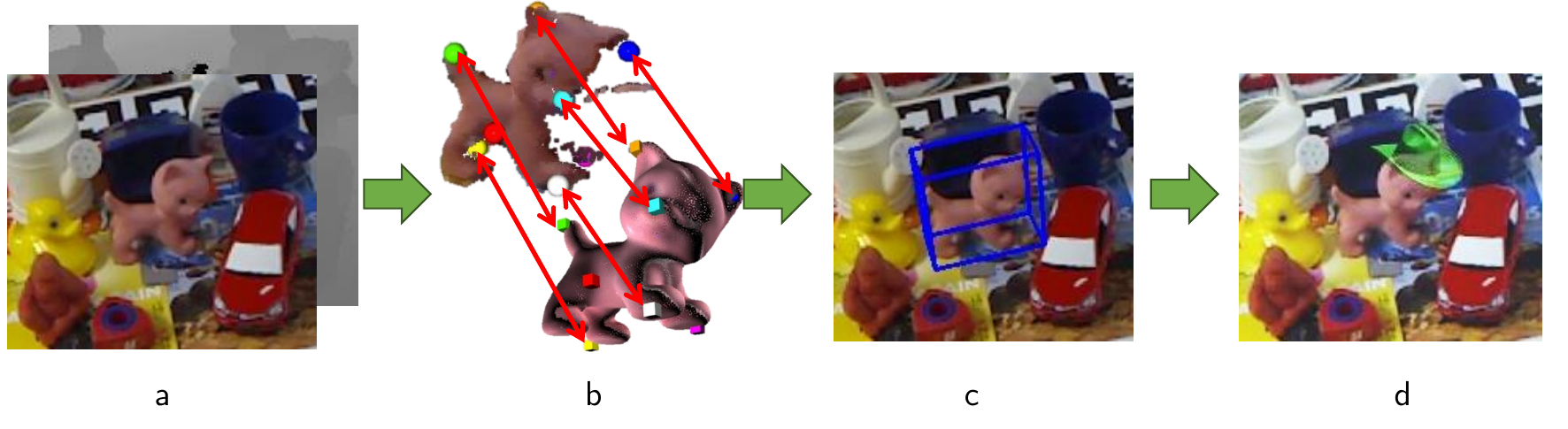}
    \caption{To estimate the pose of an object we propose to use a single RGB-D image (a) to predict the position of a set of sparse 3D keypoints (shown as spheres in b) in the camera coordinate system, which are then registered in 3D with corresponding points in the world coordinate system (shown as cubes in b) to retrieve the pose (c) that can be used to insert a virtual object (d) in the scene as an AR application}
    \label{fig:ra_examples}
\end{figure}

The goal of object pose estimation is to predict the rotation and position of an object relative to a known coordinate frame (usually the camera coordinate frame). This computer vision problem has many applications such as augmented reality or robotics. In the former case, it allows a realistic insertion of virtual objects in an image as described in \cite{marchand2015pose} and shown in Fig. \ref{fig:ra_examples}. In the latter it can be used as an input for a robotic arm to grasp and manipulate the object such as in \cite{wang2019densefusion}. Although heavily studied, this problem is still relevant as it is unresolved due to its complexity. Indeed some scenes can be highly  challenging due to the presence of cluttering, occlusions, changes in illumination, viewpoint, and textureless objects.
Nowadays color and depth (RGB-D) sensors are smaller and cheaper than ever, making them relevant for object pose estimation. Indeed, compared to color-only (RGB) sensors, the depth channel provides relevant information for estimating the pose of textureless objects in dimly lit environments.

Classical object pose estimation approaches are either based on local descriptors followed by 2D-3D correspondences \cite{marchand2015pose}, or on template matching \cite{tejani2014latent,hinterstoisser2012model,kehl2016deep}. However the challenging cases listed above limit their performance. 
To address these limitations most recent methods solve the problem of object pose estimation with a data driven strategy using for example Convolutional Neural Networks (CNNs) \cite{wang2019densefusion,rad2017bb8,peng2019pvnet,tekin2018real,li2019cdpn,park2019pix2pose,li2018deepim,xiang2017posecnn,tekin2019h+,he2020pvn3d}. These approaches work in a holistic way, considering a whole RGB or RGB-D image as an input and making a single estimation of the pose. While some methods are hybrid, using a learning-based approach followed by a geometrical solver \cite{rad2017bb8,peng2019pvnet,tekin2018real,li2019cdpn,park2019pix2pose, he2020pvn3d}, others use an end-to-end CNN to predict the pose \cite{wang2019densefusion,li2018deepim,xiang2017posecnn}.
\par
Some older methods however have proven to be reliable using patch voting approaches coupled to a learning algorithm \cite{tejani2014latent,kehl2016deep,riegler2013hough,fanelli2011real,kacete2016real}. Those strategies predict a set of pose hypothesis from local patches using data driven functions and, from this set of hypothesis, retrieve a pose.
\par
We argue that we can leverage the robustness brought by local approaches with a two stages strategy, predicting the pose in an intermediate Euclidean 3D space and retrieving it with a geometrical solver.
The intermediate representation makes it natural to apply a voting strategy to the set of pose hypothesis. Our hybrid strategy allows us to correctly supervise our CNN training, not being dependent on the choice of pose representation, not requiring a custom loss function to compute the pose error and not having to predict rotation and translation separately. Moreover, like \cite{tekin2019h+} we argue that predicting keypoints in 3D and solving a 3D-3D correspondence problem yields to more accurate results rather than predicting in 2D and solving a 2D-3D correspondence problem. 
\par
In this paper we tackle the problem of pose estimation considering a single RGB-D image as input. We design a robust and accurate algorithm to predict the pose of a generic rigid object in a scene. Our contributions are :
\begin{itemize}
	\item We propose an hybrid pipeline in two parts: a data driven block that predicts a set of 3D points in the camera coordinate system and a geometrical block. The latter retrieves the pose given the estimated points and a priori chosen keypoints in the world coordinate system, minimizing a registration error.
	\item We propose to use two CNNs in cascade in the former part. First we predict \modif{the 2D location of the object in the image}, classifying local patches extracted from the image with a CNN. Then we use a regression CNN to predict a set of possible 3D positions of points in the camera coordinate system. The position hypothesis are then robustly aggregated to obtain a single estimation of \modif{the 3D location of the points}. 
	\item We demonstrate performance improvements in terms of accuracy over state-of-the-art methods of RGB-D pose estimation on the standard LineMod \cite{hinterstoisser2012model} dataset and study the impact of some parameters of our method. We also validate our approach within a visual servoing experiment.
	\end{itemize}

\section{Close Work}
We will limit ourselves to learning based methods as the literature on object pose estimation is vast. 
We can separate those methods into two main categories: patch-based methods and holistic methods. The latter can be as well separated into two categories: direct and indirect strategies. 

~\\
\textbf{Patch-Based Methods.}
Patch-based methods output multiple pose hypothesis for a single image \cite{tejani2014latent,kehl2016deep,riegler2013hough,fanelli2011real,gall2011hough}. The predictions, called votes, which are obtained from local patches in the image are then aggregated to get a single estimation, which is more robust than each vote taken independently. Hough based methods is such a type of voting scheme. Hough Random Forests (HRFs) have been introduced by \cite{gall2011hough} to estimate the Hough transform with a learning based approach for object detection, tracking in 2D and actions recognition. The concept of HRFs has also been applied to object pose estimation by \cite{fanelli2011real} to predict the translation and rotation of human heads. In that case, both the nose 3D position and Euler angles are regressed. Those methods rely on binary tests to describe the split hypothesis used in random forests. \cite{tejani2014latent} proposes to use a split function based on a template patch descriptor. It also proposes to train a random forest using only object patches. 
As HRFs are based on handcrafted split functions, their performance is limited by image variations. To overcome this, Hough Convolutional Neural Networks (HCNNs) have been introduced by \cite{riegler2013hough} as an alternative to HRFs. A CNN was designed by \cite{riegler2013hough} to regress at once the probability of a patch belonging to the foreground as well as the object pose. 
In all cases a non parametric clustering algorithm is then used on object patches to robustly retrieve the pose. 
~\\
\textbf{Direct Holistic Methods. }
Recently, most studies \cite{wang2019densefusion,li2018deepim,xiang2017posecnn,do2018deep,kehl2017ssd} take a whole image as an input and try to leverage the capabilities of CNNs by directly estimating the pose. PoseCNN \cite{xiang2017posecnn} proposes an end-to-end CNN to perform 3 related tasks: semantic labeling, translation prediction from the object estimated 2D center and depth and rotation inference. To correctly supervise the network training, \cite{xiang2017posecnn} uses a specific loss called PoseLoss, defining the error as an average euclidean distance between rotated point clouds. SSD6D, \cite{kehl2017ssd}, uses a CNN to predict the object class with its bounding box, as well as to classify discretized viewpoints and in-plane rotations to create a set of pose hypothesis. Thus, the network loss is a parametric combination of multiple losses. 
DenseFusion, \cite{wang2019densefusion}, combines color and depth channels in a deep network to fuse them, creating a set of features which are then used by a CNN to predict the pose. It can be further rapidly refined by a network in an iterative manner. In some recent works \cite{do2018deep,sundermeyer2018implicit,mahendran20173d} the choice of representation for rotations has been studied as it shows to have an impact on the accuracy of the pose estimation \cite{mahendran20173d}. 
\\
\textbf{Indirect Holistic Methods.}
On the other side some methods \cite{rad2017bb8,peng2019pvnet,tekin2018real,li2019cdpn,pavlakos20176, he2020pvn3d, wu2016single, grabner20183d} are inspired by classical pose estimation from 2D-3D correspondence. However CNNs are used to address the limits imposed by handcrafted features. To do so the 2D location of the projection of prior chosen 3D keypoints is predicted in the image. The pose is then retrieved using a 2D-3D geometrical solver e.g. a Perspective-n-Points (PnP) algorithm.  For example BB8, \cite{rad2017bb8} coarsely segments the object and apply a deep convolutional network to the local window around the object to predict the 2D location of the projection of the 8 corners of the object bounding box. This estimation is followed by a PnP that recovers the pose. \cite{tekin2018real} proposes a single-shot CNN that classifies the object, predicts a confidence term as well as the 2D location of the projection of 9 keypoints in the bounding box. 
PVNet\cite{peng2019pvnet} proposes to apply an offset based approach to predict the 2D location of a set of keypoints on the object surface. To do so, they segment the object in the image and predict a vector field on the segmented object, the spatial probability distribution of each keypoint is then retrieved and used in an uncertainty driven PnP to estimate the pose. H+O \cite{tekin2019h+} estimates at once hand-object poses as well as objects and actions classes from RGB images. A CNN predicts the 3D position of 21 points of the object bounding box and the object pose can be retrieved from 3D-3D correspondences. \modif{PVN3D \cite{he2020pvn3d} uses 6 different networks to estimate the pose: 3 (including \cite{wang2019densefusion}) are used to extract and fuse features from a single RGB-D image and 3 are used to predict a set of 3D keypoints and the objects segmentation.}

\section{Proposed Approach: our Hybrid, Patch-Based Strategy}

\begin{figure*}
    \centering
    \includegraphics[scale=0.32]{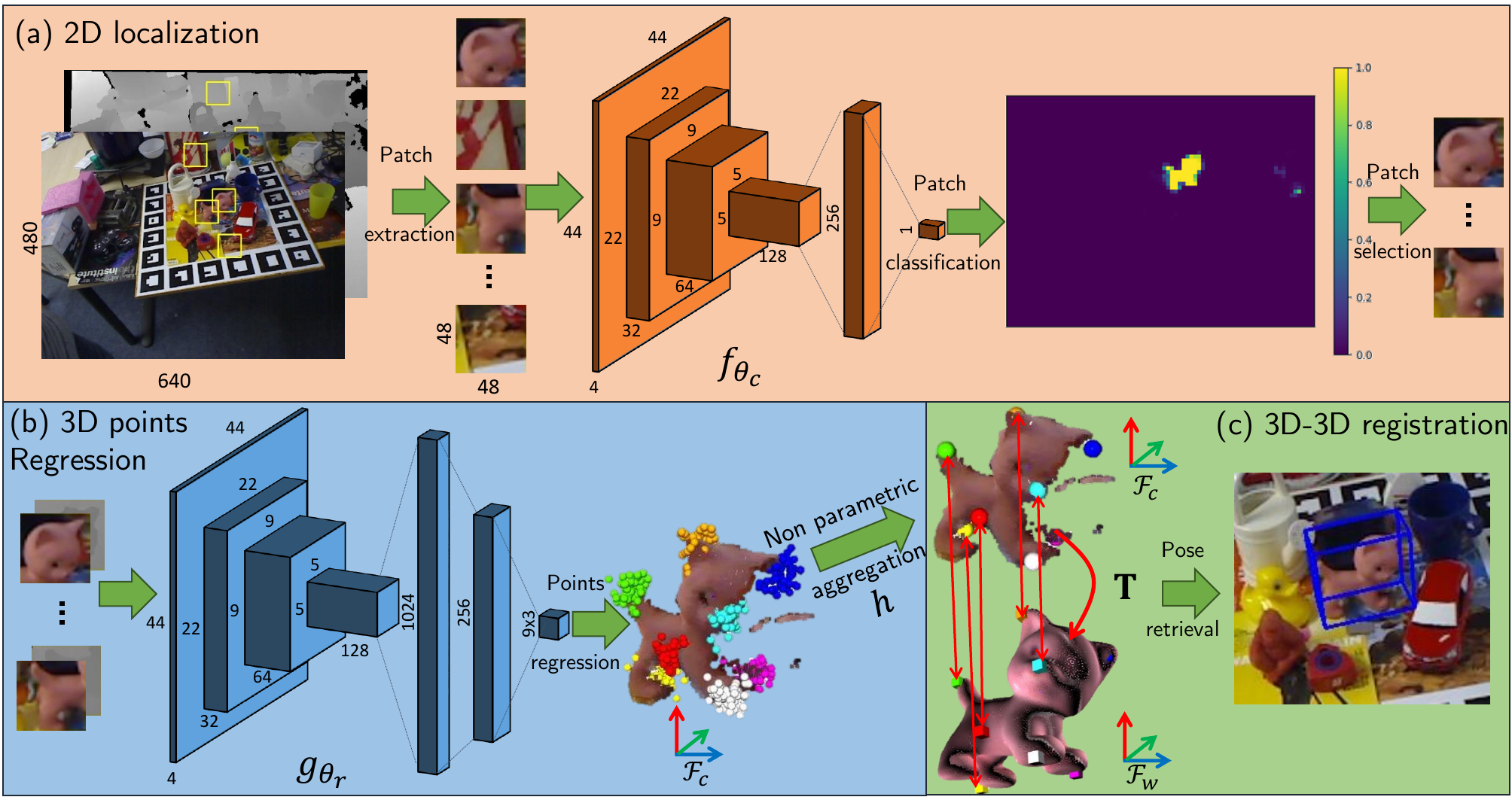}
    \caption{Overview of our pipeline solution: first (a), patches are extracted from an RGB-D image and classified as object or background. Next (b), for each object patch, a regression CNN predicts a set of vectors estimating the position of specific 3D keypoints in $\mathcal{F}_c$. Those votes are then aggregated in a non parametric way to obtain a robust estimation of the points in $\mathcal{F}_c$. (c) by minimizing the registration error between corresponding estimated keypoints in $\mathcal{F}_c$ and reference keypoints in $\mathcal{F}_w$ we retrieve the 6D pose. The pipeline is illustrated here with patches of size $48\times 48$.}
    \label{fig:main_fig}
\end{figure*}
Our goal is to achieve a robust and accurate 6-DoF pose estimation of a 3D object, i.e. to estimate the transformation $\mathbf{T} \in SE(3)$ of an object of interest, provided with a coordinate system called world coordinate system $\mathcal{F}_w$, in the camera coordinate system $\mathcal{F}_c$. We represent the transformation $\mathbf{T}$ between $\mathcal{F}_c$ and $\mathcal{F}_w$ as 
\begin{equation}
\mathbf{T} = \begin{pmatrix}
\mathbf{R} & \mathbf{t}\\
\mathbf{0} & 1
\end{pmatrix}
\label{eq:}
\end{equation}
where $\mathbf{R} \in SO(3)$ is a 3D rotation matrix and $\mathbf{t} \in \mathbb{R}^3$ is a 3D translation vector.
\par
The pipeline of the proposed strategy can be seen Fig. \ref{fig:main_fig}: we adopt a patch voting based approach inspired from \cite{riegler2013hough,gall2011hough}, using multiple local information to predict a sparsified version of the object geometry in $\mathcal{F}_c$. First, we design and train a classification CNN to predict the class of patches extracted from the input image either as object or background. This allows us to roughly localize the object in 2D. Then, we design and train a regression network to predict for each extracted object patch the 3D position in $\mathcal{F}_c$ of a set of prior chosen keypoints, selected in $\mathcal{F}_w$. Finally, by minimizing the 3D-3D registration error between corresponding estimated keypoints in $\mathcal{F}_c$ and reference keypoints in $\mathcal{F}_w$ we retrieve the 6D pose. \modif{Hence our method belongs to both the indirect and patch-based set of methods.}

\subsection{2D Localization} \label{sec:2dloc}
In this section we show how we take in account the visibility of the object in each patch. Indeed not all patches contain relevant information about the object pose. 
Unlike in \cite{riegler2013hough} who uses a single network for both classification and pose estimation, we first use a classification network to decide whether or not a patch contains a representation of the object. We argue that classifying the patches, keeping only relevant ones before transmitting them to the regression network allows the CNN to fit using only relevant information about the object pose. Moreover we do not need a sophisticated parametric loss function whose parameters have to be optimized to supervise the training.
~\\
~\\
\textbf{Model.}
Our model is inspired by a light VGG-like architecture and can be seen in the first block of Fig. \ref{fig:main_fig}. It is composed of a set of convolutional layers to extract features from the images and max-pooling layers to introduce scale invariance followed by 2 dense layers to classify the extracted features. For the last layer, we use a sigmoid activation function, for each other layer we use the classical ReLu activation function. 
To help reduce overfitting dropout is also used on the first fully connected layer as it contains the most weights.
~\\
~\\
\textbf{Data.}
To train our classification network in a supervised manner, we need labeled data. We capture a set of images representing the object of interest from multiple points of view. The classification neural network is trained using a set of patches $\left\{P_i = (\mathbf{I}_i, b_i)\right\}$ where $\mathbf{I}_i$ is the RGB image of the patch of size $[h \times w]$, i.e. $\mathbf{I}_i \in \mathbb{R}^{[h \times w] \times 3}$ and $b_i \in [0, 1]$ represents whether or not the object is visible in the image $\mathbf{I}_i$. We obtain it by producing a binary mask of the object created by a 2D projection of the object 3D model using its ground truth pose. To increase the robustness of our algorithm across changes in illumination we proceed to do data augmentation by randomly modifying patches brightness.
~\\
~\\
\textbf{Training.}
 We denote the classification function $f_{\theta_c}$ optimized over $\theta_c$  which represents our CNN weights. The classification parameters are optimized by minimizing over the training data set:
\begin{equation}
\theta^*_c = \argmin_{\theta_c} \mathcal{L}_c(b, \hat{b}).
\label{eq:argmin_classif}
\end{equation}
where $\hat{b} = f_{\theta_c}(\mathbf{I})$ and $\mathcal{L}_c$ is the classical weighted binary cross entropy. 

\begin{figure}[ht]
    \centering
    \includegraphics[width=1.0\columnwidth]{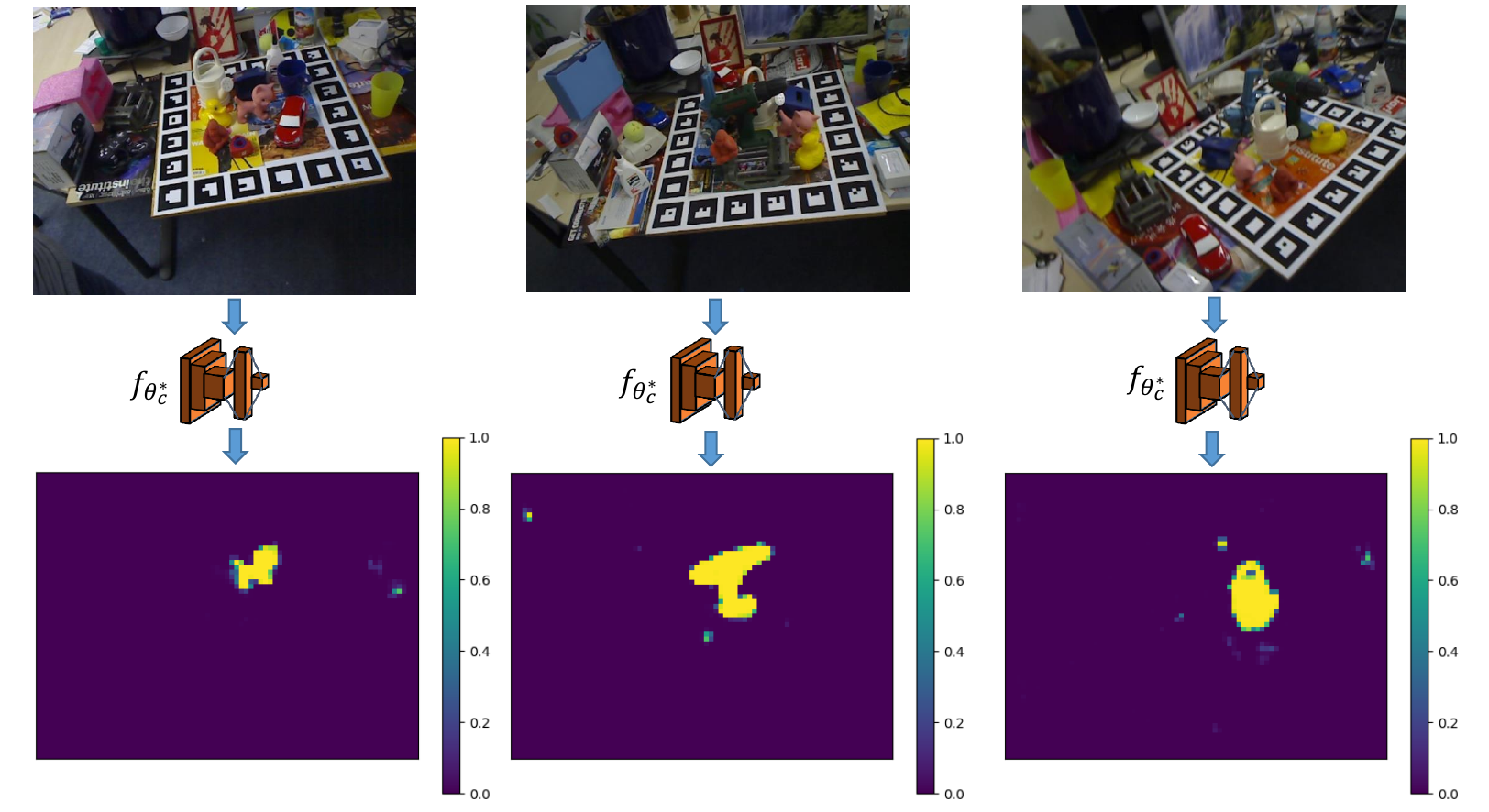}
    \caption{Examples of probability maps for the cat, driller and can from the LineMod dataset}
    \label{fig:classif}
\end{figure}

~\\
\textbf{Inference.}
Given an unseen image, we \modif{extract $K$ patches from the image in a sliding window fashion, with $K$ depending on the window stride $d$ (in our experiment $K$ can vary from to a few hundreds to a few thousands with $d$ going from 4 to 48 pixels)} and get a set of patches $\mathcal{P} = \left\{P_i, i \in [1, K]\right\}$. Each patch is then fed to the classification network $f_{\theta^*_c}$ whose output is $\left\{\hat{b}_i = p(b_i| \mathbf{I}_i) = f_{\theta^*_c}(\mathbf{I}_i), i \in [1,K]\right\}$ where $p$ denotes the probability. 
We show in Fig. \ref{fig:classif} some heat maps obtained using the probability estimated for each patch. We can see that the patches extracted from the object have a high probability of being classified as object while the patches extracted from the background have a low probability. 
\modif{As we can see the probability maps are very similar to 2D segmentation. However our method gives coarser results and needs less data than classical segmentation methods. Moreover our strategy is flexible as we can tune the patch extraction stride to balance inference time and segmentation precision.}

\subsection{3D Points Prediction}
We now show how we predict the position of a set of 3D keypoints in $\mathcal{F}_c$, using the object patches classified in the previous step. 
We use a regression network to predict the 3D location of a set of $M$ points in $\mathcal{F}_c$. First, \modif{using a point selection algorithm like the farthest point sampling algorithm} we create a set of $M$ 3D keypoints, denoted $\mathcal{S} = \left\{\mathbf{X}_j \in \mathbb{R}^3, j \in [1, M]\right\}$, chosen in the object model in $\mathcal{F}_w$. For a given pose $\mathbf{T}$ of the object in $\mathcal{F}_c$ we express the points in $\mathcal{S}$ in $\mathcal{F}_c$, and denote the set $\mathcal{S}_{\mathbf{T}} \coloneqq \left\{\mathbf{Y}_j \coloneqq \mathbf{T} \mathbf{X}_j, j \in [1, M]\right\}$. Our goal is to estimate the location of the points in $\mathcal{S}_{\mathbf{T}}$ i.e. to estimate the location of the keypoints of $\mathcal{S}$ in $\mathcal{F}_c$. We argue that it is easier for the neural network to predict points in a euclidean space than to predict a pose over $SE(3)$. Let us recall that no distance exists over $SE(3)$ which makes a loss function very difficult to exhibit. Like \cite{li2019cdpn} we argue that rotations and translations should be treated differently or at least that adaptation is required to learn  to regress coherently in $SE(3)$. In a way with our change of variables we suppress the direct impact of the peculiarities of rotation space as every variable stays in $\mathbb{R}^3$. Furthermore we argue that establishing 2D-3D correspondences is more ambiguous than 3D-3D correspondences for object pose estimation. Indeed the PnP algorithm aims at minimizing the 2D projection errors of keypoints. However, there may be small reprojection errors between keypoints that are large in 3D. \modif{Comparisons between 2D-3D and 3D-3D correspondences can be found in \cite{he2020pvn3d}.}

\begin{figure}[h]
    \centering
    \includegraphics[width=1.0\columnwidth]{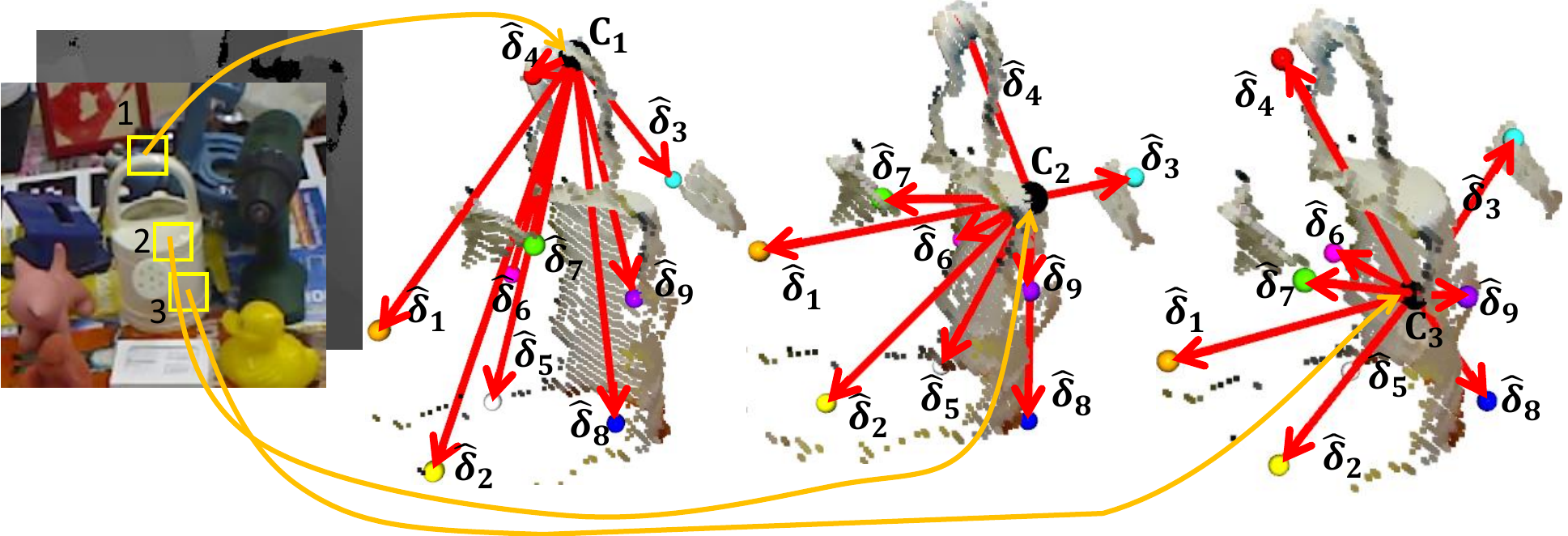}
    \caption{Example of 3 patches extracted from an RGB-D image with the estimated offsets for each patch in red and the corresponding points represented by colored spheres, the center of each patch is shown by a black sphere}
    \label{fig:offsets}
\end{figure}

~\\
\textbf{Model.}
The architecture of the regression network can be seen in the second block of Fig. \ref{fig:main_fig}.
We use an architecture that is very close to the classification network because we showed that we could reliably extract information from the patches with it. However we change the fully connected part, adding one layer and using more weights for each layer to give the regression CNN more flexibility. 
\newline
\textbf{Data.}
We extract only object patches $P_i'$ from the image. A regression neural network is trained using a set of patches $ \mathcal{P}' = \left\{P'_i = (\mathbf{I'}_i, \bm{\delta}_i)\right\}$ where $\mathbf{I'}_i$ is the RGB-D image of the patch, i.e., $\mathbf{I'}_i \in \mathbb{R}^{[h \times w] \times 4}$ and $\bm{\delta}_i \in \mathbb{R}^{3 \times M}$ is a set of $M$ 3D vectors, called offsets and defined in Equation \ref{eq:offset}: 
\begin{equation}
\begin{split}
\bm{\delta}_i & =\left\{\bm{\delta}_{1,i}, \bm{\delta}_{2,i}, ..., \bm{\delta}_{M,i}\right\} \\ 
& = \left\{\mathbf{Y}_1-\mathbf{C}_i, \mathbf{Y}_2 - \mathbf{C}_i, ..., \mathbf{Y}_M-\mathbf{C}_i\right\}
\end{split}
\label{eq:offset}
\end{equation}
with $\mathbf{Y}_j \in \mathcal{S}_{\mathbf{T}_i} \quad \forall j \in [1,M]$, $\mathbf{T}_i$ is the pose of the object visible in the $i^{th}$ patch and $\mathbf{C}_i \in \mathbb{R}^3$ is defined by :
\begin{equation}
\mathbf{C}_i = 
\begin{pmatrix}
\frac{u_i-c_x}{f_x}Z_i\\
\frac{v_i-c_y}{f_y}Z_i \\
Z_i
\end{pmatrix}
\label{eq:back_pinhole}
\end{equation}
with $f_x$, $f_y$, $c_x$, $c_y$ the camera intrinsics, $(u_i,v_i)$ the 2D position of the center of the $i^{th}$ patch and $Z_i$ the value of the patch depth at location $(u_i, v_i)$.
Equation \ref{eq:back_pinhole} corresponds to the 3D backprojection of the 2D center of the $i^{th}$ patch, using a pinhole model.
Thus, $\bm{\delta}_i$ is a set of $M$ vectors, each one going from the 3D center of the patch and one of the $M$ points in $\mathcal{S}_{\mathbf{T}_i}$. An example of offsets is visible in Fig. \ref{fig:offsets}: for 3 patches extracted in the image, we show $M = 9$ offsets. 
The use of offsets is very interesting for object pose estimation for two reasons. First, offsets bring invariance translation that is necessary due to the fact that we consider local patches. Indeed, 2 patches extracted from 2 different images with different poses may be very resembling. If displacement vectors are not used, the difference in terms of pose can thus only be seen as noise by the network. On the contrary, if offsets are employed the variable to regress is more correlated to patches aspect. Second, let's consider the space of all possibles object translation denoted $\Omega_t$, if we do not use offsets then this space is at most $\mathbb{R}^3$. However when considering displacement vectors, the set of all possible offsets has an upper bound of $D$ where $D$ is the largest diameter of the object, thus $\Omega_t \subseteq \Omega_{\delta} = B(0,D)$ where $B(0,D)$ is the ball of center $0$ and radius $D$ and necessarily we have $\Omega_{\delta} \subset\mathbb{R}^3$. \modif{The manifold in which predictions exist is reduced when using offsets, making the learning procedure easier for a data driven algorithm.}

\textbf{Training.}
We denote the regression function $g_{\theta_r}$ where $\theta_r$ is the vector of weights of the network. The regression parameters are optimized by minimizing over the training data set:
\begin{equation}
\theta^*_r = \argmin_{\theta_r} \mathcal{L}_r(\bm{\delta}, \hat{\bm{\delta}})
\label{eq:argmin_reg}
\end{equation}
where $\hat{\bm{\delta}} = g_{\theta_r}(\mathbf{I})$ and $\mathcal{L}_r$ is the mean absolute error: 
\begin{equation}
\mathcal{L}_r(\bm{\delta}, \hat{\bm{\delta}}) = \frac{1}{3M}||\bm{\delta}-\hat{\bm{\delta}}||_1
\label{eq:loss_reg}
\end{equation}
where $||.||_1$ is the usual $L^1$ norm, thus $\mathcal{L}_r$ represents the averaged $L^1$ distance between the estimated and ground truth points. The $L^1$ norm is preferred to the $L^2$ norm because it is less sensitive to outliers, that are robustly handled \modif{by the Gaussian kernel of the mean-shift algorithm (see below) during the voting step.}

\begin{figure}[h]
    \centering
    \includegraphics[width=1.0\columnwidth]{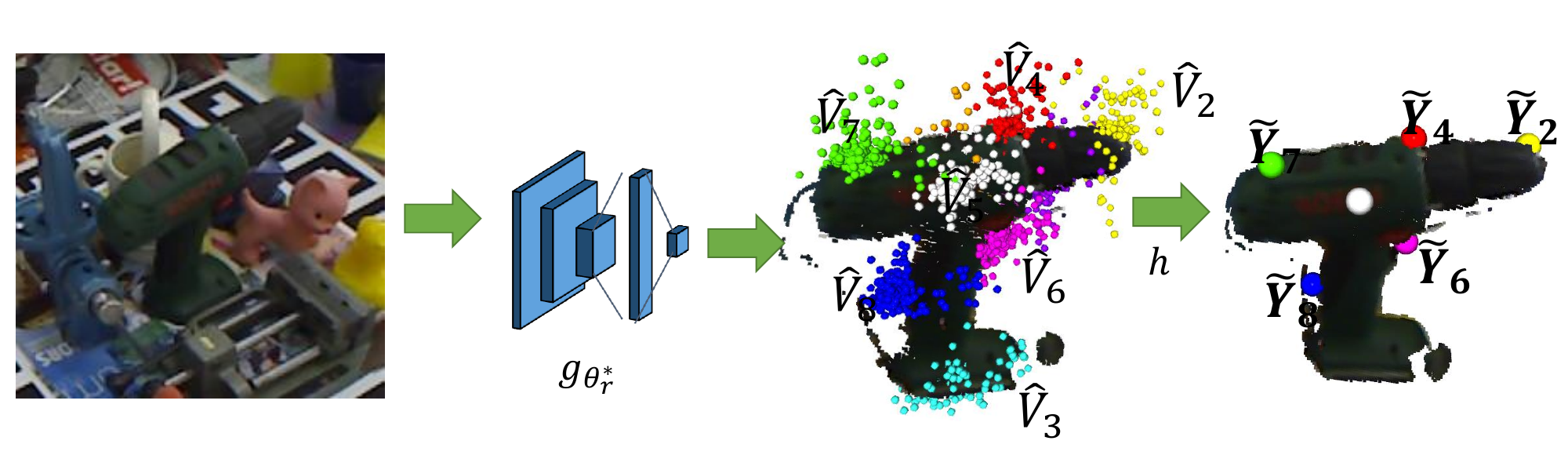}
    \caption{Example of predicted 3D points. From left to right: the cropped input image, the cluster of 3D votes $\hat{V}_j $where each color corresponds to votes for a single point, the aggregated points $\tilde{\mathbf{Y}}_j$ obtained by mean-shift (best seen in color). The aggregated point of cluster $\tilde{V}_3$ is hidden behind the driller point cloud.}
    \label{fig:votes_examples}
\end{figure}

\textbf{Inference and Voting.}
Given the patches extracted in Sec. \ref{sec:2dloc} and their associated estimated probability, we discard the patches which probability is lower than a threshold $\tau$. Thus we get a set: $\mathcal{P}_{\tau} = \left\{\mathbf{I}_i | \hat{b}_i > \tau , i \in [1,K] \right\}$ that can be written: $\mathcal{P}_{\tau} = \left\{\mathbf{I}_i, i \in [1,N] \right\}$ \modif{where $N$ is the number of object patches that can go from a few tens to a few hundreds}.
For the $i^{th}$ patch $\mathbf{I}_i$ fed to the regression network we get $M$ predicted 3D offsets $g_{\theta_r^*}(\mathbf{I}_i)$ denoted $\hat{\bm{\delta}_i} = \left\{\hat{\bm{\delta}}_{j,i}, j \in [1,M] \right\}$. We can then get an estimation of the 3D location of the transformed points by adding the position $\mathbf{C}_i$ of the 3D center of the patch obtained from \ref{eq:back_pinhole}. This way we get a set of $M$ estimated points positions $\left\{\hat{\mathbf{Y}}_{j, i}, j \in [1,M] \right\}$ in $\mathcal{F}_c$. When we take in account all the $N$ patches we get $N \times M$ points: $\hat{V} \coloneqq \left\{\hat{\mathbf{Y}}_{j,i}, i \in [1, N], j \in [1,M] \right\}$ that can be viewed as $M$ clusters of $N$ points or votes in the $\mathcal{F}_c$. We denote the $j^{th}$ cluster of points $\hat{V}_j \coloneqq \left\{\hat{\mathbf{Y}}_{j,i}, i \in [1, N] \right\} $. The votes must then be aggregated to get a robust estimate of the 3D position of each point in the $\mathcal{F}_c$. We denote the aggregation function $h : \mathbb{R}^{3 \times N} \longmapsto \mathbb{R}^3$. It is necessary to aggregate the $N$ 3D votes in a robust manner to limit the impact of possible outliers, hence $h$ is chosen to be a non-parametric estimator of the maxima of density. In our case we use a mean-shift estimator \cite{comaniciu2002mean,cheng1995mean} which iteratively estimates the local weighted mean in \ref{eq:mean_shift}: 
\begin{equation}
m(\mathbf{X}) = \frac{\sum_{i}{k(\mathbf{X}_i-\mathbf{X})}\mathbf{X}_i}{\sum_{i}{k(\mathbf{X}_i-\mathbf{X})}}
\label{eq:mean_shift}
\end{equation}

where $k$ is a kernel function such as a Gaussian kernel: $k_{\sigma}(\mathbf{X} -\mathbf{Y}) = \exp(-\frac{||\mathbf{X}-\mathbf{Y}||^2}{2\sigma^2})$ \modif{which gives less weight to outliers}. \modif{The parameter $\sigma$ was not optimized here but it can simply be chosen to optimize a pose error metric on a validation set.}
Thus we can define the set $\hat{\mathcal{S}_T} \coloneqq  \left\{\tilde{\mathbf{Y}}_j \coloneqq h(\hat{V}_j), \forall j \in [1, M] \right\}$ which corresponds to the aggregated centroid of each cluster in $\mathcal{F}_c$. We show Fig. \ref{fig:votes_examples} such examples of votes. 

\subsection{3D-3D Correspondence Alignment}
In this section we show how to retrieve the pose using the estimated 3D keypoints in $\mathcal{F}_c$ that we obtained in the previous step and their corresponding reference keypoints in $\mathcal{F}_w$. 
Once the centroids have been voted we align the estimated points and their corresponding reference to get a pose estimation from the estimated location of the points. To do so we seek to find the transformation $\mathbf{T}^* \in SE(3)$ that minimizes: 
\begin{equation}
\mathbf{T}^* = \argmin_{\mathbf{T}}\sum_{j=1}^M{||\mathbf{T}\mathbf{X}_j-\tilde{\mathbf{Y}}_j||^2_2}
\label{eq:3d_solver}
\end{equation}
where $\mathbf{T}$ can also be represented with a minimal representation $q \in \mathfrak{se}(3)$ where $\mathfrak{se}(3)$ is the Lie algebra associated to $SE(3)$, $X_j \in \mathcal{S}$, $\tilde{Y_j} \in \hat{\mathcal{S}_T}$ and $||.||_2$ is the euclidean norm of $\mathbb{R}^3$. That is finding the pose that best fits the points estimated by the aggregation of votes in $\mathcal{F}_c$. 
This problem is called the Orthogonal Procrustes Problem and can be solved using SVD decomposition as shown in \cite{arun1987least} or an Iteratively Reweighted Least Square algorithm \cite{fitzgibbon2003robust,malis2006experiments} to discard outliers and obtain a robust estimation.
To further refine the pose we can apply an Iterative Closest Point (ICP) algorithm \cite{besl1992method}. This consists in solving \ref{eq:3d_solver}, using the 3D model points and the points measured by the RGB-D camera projected in 3D using \ref{eq:back_pinhole}.

\section{Experiments}
We now present the results we obtained on the LineMod \cite{hinterstoisser2012model} dataset. This section is divided in four parts: first we present the technical details of our implementation, then we evaluate our method in terms of classification accuracy, 3D points regression accuracy and we measure the object pose accuracy using a standard metric and compare it to state-of-the-art results. Finally we compute the average inference time for a given object and study the impact of some hyper-parameters such as the level of patch density on both pose accuracy and inference time. Last but not least, a visual servoing task based on the proposed method is also considered.
\subsection{Implementation Details}

\textbf{Training.}
To build our training data we extract patches in a sliding window fashion. 
We train the classification network for 100 epochs and the regression network for 500 epochs. We use a learning rate of $10^{-4}$ with the Adam optimizer. 
A dropout of 50\% is used for the classification CNN and and dropout of 20\% and 10\% is used on the two first fully connected layers of the regression CNN.
We implement the neural networks using the tensorflow \cite{tensorflow2015-whitepaper} framework. 
\newline
~\
\textbf{Keypoints Selection.}
Inspired by \cite{peng2019pvnet}, we select the keypoints using the farthest point sampling algorithm which allows us to get a good coverage of the object. In our experiments we chose to use \modif{M = } 9 points. 
\newline
~\
\textbf{Inference.} 
Our algorithm is implemented in Python. During the inference we extract patches with a stride $d$ of 4 pixels. We chose to set the threshold $\tau$ at 0.98.
The votes are aggregated using the mean-shift algorithm and a gaussian kernel with variance $\sigma^2 = 40^2 mm$. 
We use open3d ICP, on the sub map defined by the estimated bounding box. For testing we use a Nvidia RTX2070 and an Intel Xeon @3.7 GHz. \modif{The patch size is chosen to be $h\times w = 64\times64$ unless precised otherwise.}
\subsection{Datasets}The LineMod dataset consists of about 15 000 RGBD images of 13 objects with multiple recorded poses in a variable and cluttered environment. It is widely used by the pose estimation community. We use the same method as \cite{wang2019densefusion,rad2017bb8,peng2019pvnet} to select training and testing images.
The dataset being small (about 100 images for training), it is very challenging for CNN based approaches as highlighted in \cite{xiang2017posecnn}. This makes some methods like \cite{peng2019pvnet,xiang2017posecnn} need synthetic data. 
\modif{Very large datasets such as the YCB-Video dataset \cite{xiang2017posecnn} may not be available for some objects as their creation is expensive in time as well as complex.}
\modif{Using data augmentation on the brightness we manage to get state of the art results without needing additional synthetic data, contrary to \cite{he2020pvn3d, peng2019pvnet} that use 20 000 new synthetic images per object.}

\subsection{Classification Accuracy}
In this subsection, we measure the performance of the classification network. 
Having a bad classification accuracy could lead to multiple patches being misclassified. A high false positive rate would create noise in the Hough space and complexify the task of finding the maximum density. On the contrary, a high false negative rate would reduce the number of patches used for regression and thus the number of votes, leading to a less robust estimation. 
We can see in table \ref{table:accuracy_class} that for every object we get a high true negative rate above 99.5 \% meaning we do not pollute the vote space. The true positive rate is more variable but stays above about 90\%, so not too many patches are discarded. 
\begin{table}
\centering
\caption{True positive rate \modif{(TPR)} and true negative rate \modif{(TNR)} (in \%) for each object using our classification network}

\begin{tabular}{|c|c|c|c|c|c|c|c|}
\hline
 & \multicolumn{1}{|l|}{ape} & \multicolumn{1}{|l|}{ben.} & \multicolumn{1}{|l|}{cam} & \multicolumn{1}{|l|}{can} & \multicolumn{1}{|l|}{cat} & \multicolumn{1}{|l|}{drill.} & \multicolumn{1}{|l|}{duck} \\ \hline
\modif{TPR} & 98.2 & 89.8 & 92.8 & 91.0 & 97.5 & 95.1 & 91.3  \\ 
\modif{TNR} & 99.9 & 99.5 & 99.6 & 99.7 & 99.7 & 99.7 & 99.9 \\ \hline
& \multicolumn{1}{|l}{hole.} & \multicolumn{1}{|l|}{iron} & \multicolumn{1}{|l|}{lamp} & \multicolumn{1}{|l|}{eggbox} &\multicolumn{1}{|l|}{glue} & phone & MEAN \\ \hline
\modif{TPR} & 97.1 & 94.9 & 88.4 & 94.2 & 90.7 & 89.6 & 93.1\\
\modif{TNR} & 99.7 & 99.5 & 99.5 & 99.8 & 99.8 & 99.5 & 99.7  \\ \hline
\end{tabular}
\label{table:accuracy_class}
\end{table}
\subsection{3D Points Regression Accuracy}
In this subsection, we study the accuracy of the regression network. For each object we measure the average euclidean distance between the estimated position of each keypoint after it has been aggregated and its ground truth position


\begin{table}
\caption{Average euclidean distance and standard deviation (in mm) between ground truth and predicted 3D points for each object}
\centering
\begin{tabular}{|c|c|c|c|c|c|c|c|}
\hline
 & \multicolumn{1}{|c|}{ape} & \multicolumn{1}{|c|}{ben.} & \multicolumn{1}{|c|}{cam} & \multicolumn{1}{c|}{can} & \multicolumn{1}{c|}{cat} & \multicolumn{1}{c|}{drill.} & \multicolumn{1}{c|}{duck} \\ \hline
Avg. & 7.2 &12&16.9&11.7&9.4&12.8&10.1 \\

Std. & 4.5&7&65.6&6.7&5.5&8.3&6.3\\\hline
&\multicolumn{1}{c|}{hole.} & \multicolumn{1}{c|}{iron} & \multicolumn{1}{c|}{lamp}& \multicolumn{1}{c|}{egg.}& \multicolumn{1}{c|}{glue} & phone & MEAN \\ \hline
Avg.(mm) &10.2&10.6&12.4&14&14.6&11.1& 11.8\\
Std. (mm) &19.6&36.9&42.6&8.8&14.4&22.9&19.2  \\ \hline
\end{tabular}
\vspace{-5mm}
\label{table:reg_error}
\end{table}

We can see \modif{in table \ref{table:reg_error}} that the euclidean distance between predicted and ground truth points is \modif{on average of 11.8 mm}

\subsection{Object Pose Accuracy}
\textbf{Metric.}
We use the standard 6 DoF metric developed in \cite{hinterstoisser2012model}, the average distance of model points (ADD-S).
A pose is considered correct if the value of the ADD-S is less than 10\% of the object diameter $D$.
We report the results in table \ref{table:results}. 
\modif{As we can see we obtain on average very close results to \cite{he2020pvn3d} while using about 200 times less data which shows the superiority of our method in the small data regime. For some objects we even obtain better results than \cite{he2020pvn3d}.}

\begin{table}[h]
\caption{Percentage of correctly predicted poses using the ADD-S metric on the LineMod dataset compared to state-of-the-art methods. Eggbox and glue are considered as symmetric objects.}
\center
\begin{tabular}{|l|c|c|c|c|c|l}
\cline{1-6}
\multicolumn{1}{|c|}{Input} & \multicolumn{1}{c|}{RGB} & \multicolumn{4}{c|}{RGB-D} &  \\ \cline{1-6}
Method & \multicolumn{1}{|l|}{\cite{peng2019pvnet}}& \multicolumn{1}{|l|}{\cite{wang2019densefusion} w. ref.} &\multicolumn{1}{|l|}{\cite{he2020pvn3d}} &  \multicolumn{1}{|l|}{Ours} & \multicolumn{1}{|l|}{Ours + ICP} &\\ \cline{1-6}
ape &  43.6 & 92.0 & \textbf{97.3} & 91.2&\textbf{97.3}&\\
ben. & 99.9 & 93.0 &99.7 & \textbf{100.0}&\textbf{100.0}& \\
cam &  86.9 & 94.0 & \textbf{99.6} & 95.2&98.4& \\
can & 95.5 & 86.0 & \textbf{99.5} &  98.1&\textbf{99.5}&\\
cat  & 79.3 & 93.0 & \textbf{99.8} &  98.4&99.7&\\
drill.  & 96.4 & 97.0 &99.3 &  98.8&\textbf{99.8}&\\
duck  & 52.6 & 87.0 & \textbf{98.2} &  82.2&98.0&\\
hole. & 81.9 & 92.0 &\textbf{99.8} &  93.7&98.8&\\
iron  & 98.9 & 97.0 & \textbf{100.0} & 99.1&99.9& \\
lamp & 99.3 & 95.0 & \textbf{99.9} &  98.6&99.1&\\
egg. (sym.)  & 99.2 & \textbf{99.8} & 99.7 &  99.3&99.3&\\
glue (sym.) & 95.7 & \textbf{100.0} & 99.8 &  99.2&99.0&\\
phone  & 92.4 & 93.0 &99.5 &  98.3&\textbf{99.6}&\\ \cline{1-6}
MEAN & 86.3 & \multicolumn{1}{|c|}{94.3}& \multicolumn{1}{c|}{\textbf{99.4}}  &  \multicolumn{1}{c|}{96.3} &99.1 &\\ \cline{1-6}
\end{tabular}
\label{table:results}
\end{table}

\begin{figure}
    \centering
    \includegraphics[width=1.0\columnwidth]{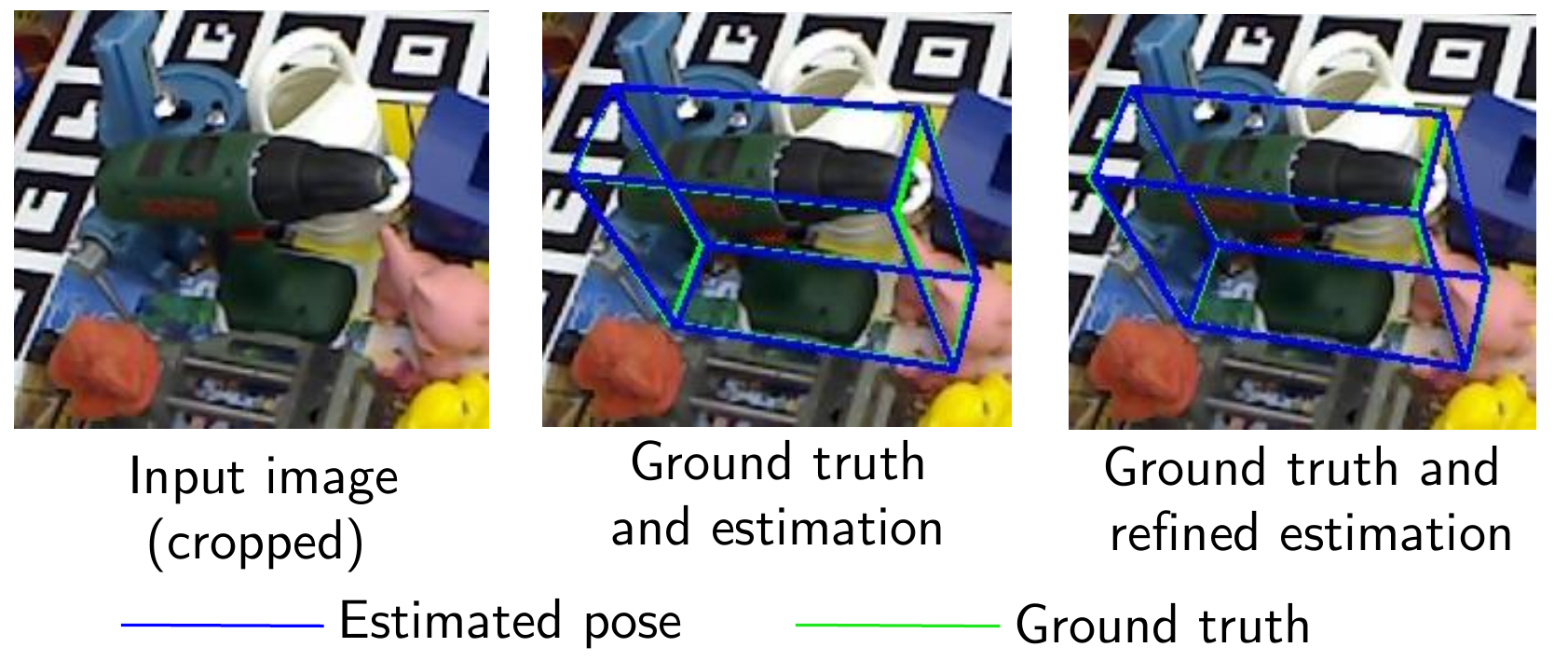}
    \caption{Some qualitative examples of pose estimation results on LineMod. The pose estimation is represented as the blue bounding box and the ground truth pose as the green bounding box}
    \label{fig:quali_results}
\end{figure}

\subsection{Inference Time}
\begin{figure}
    \centering
    \includegraphics[width=1.0\columnwidth]{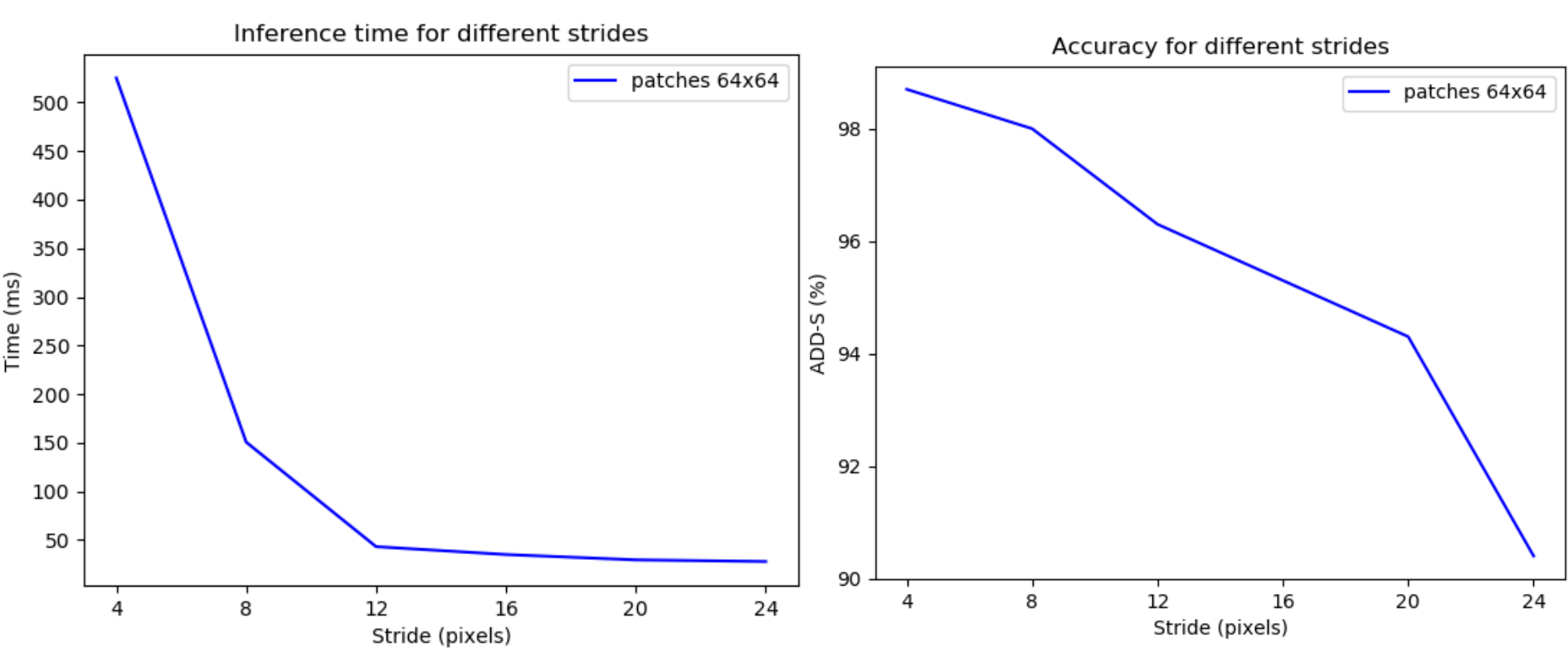}
    \caption{Inference time (in ms) and accuracy for the driller for varying strides (in pixels)}
    \label{fig:time_graphs}
\end{figure}

Inference time is greatly dependent on the choice of the density with which patches are extracted. The lower the stride is, the more patches have to be extracted and fed to the networks and the longer the inference will be. However we expect the accuracy to be growing with the number of patches extracted. This balance allows our method to be suitable to a wide range of methods. The flexibility it brings lets the user tune the extraction stride to better meet the application needs. 
\modif{To decrease inference time we retrained a light 2D detection algorithm (namely tiny YOLOv3 \cite{redmon2018yolov3} with a darknet backbone) on the driller. As the training set is very small the estimated bounding box is coarse but sufficiently precise to reduce inference time which is reported in table \ref{time_yolo}. As we can see, using a stride of 12 to 16 we can reach real time inference while losing little accuracy. To speed up the voting step we chose to use only 3 clusters that are selected to minimize their respective variance.}
\modif{We also report inference times and accuracy for different strides in \ref{fig:time_graphs}.}

\begin{table*}
\centering
\caption{Inference times (in ms) and pose accuracy (ADD-S \%) for the driller, using tiny YOLOv3 and different strides (pixels).}
\begin{tabular}{|c|c|c|c|c|c|c|c|c|}
\hline
 Stride & \multicolumn{1}{|c|}{Bbox estimation} & \multicolumn{1}{|c|}{Patch extract.} & \multicolumn{1}{|c|}{Classification} & \multicolumn{1}{|c|}{Regression.} & \multicolumn{1}{|c|}{Voting} & 3D-3D solving & Total & ADD-S \\ \hline
 4 &8.8 &	83.4&	187.1&	69.1&	176.5&	0.3&525.2& 98.7 \\
8 &9&	20.8&	51.2&	23&	46.4&	0.2&	150.6& 98\\
12 &8.7&	3&	11&	9&	11.1&	0.3&	43.1& 96.3\\
16 &8.9&	1,9&	8,6&	7.6&	7.7&	0.3&	35& 95.3\\
20 &8.6&	1.2&	7.6	&6.4	&5.4&	0,3	&29.5& 94.3\\
24 &9&	1&	7.4&	5.5&	4.6&	0.3	&27.8& 90.4\\ \hline
\end{tabular}
\label{time_yolo}
\vspace{-3mm}
\end{table*}

\subsection{Ablation studies.}
\begin{figure}
    \centering
    \includegraphics[width=0.6\columnwidth]{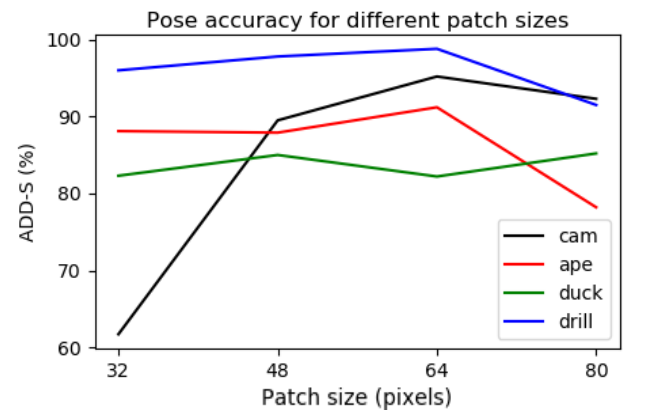}
    \caption{ADD-S (\%) for different patch size (in pixels) and objects.}
    \label{fig:accu_patch_size}
    \vspace{3mm}
\end{figure}

\modif{\textbf{Influence of patch size.}}
\modif{In this section we study the influence of different patch sizes for 4 LineMod objects on the pose estimation accuracy. Patch size is an important parameter and depends on the object size and texture. For example textureless objects may be better represented using larger patches that can capture some of the object shape.The influence on inference time can be seen for the driller in table \ref{table:size_inf_time}, as expected inference time grows with patch size as more convolutions are needed. The pose estimation accuracy can be seen in figure \ref{fig:accu_patch_size} and shows that pose accuracy grows with patch size, up to $64\times 64$ and then goes down for most objects.}

\begin{table}[]
\center
\caption{Total inference time (in ms) for each patch size (in pixels) with a stride of 4 pixels}
\begin{tabular}{|l|llll|}
\hline
 Patch size (pixels)                    & 32    & 48  & 64    & 80    \\ \hline
Total inference time (ms) & 391.6 & 413 & 525.2 & 634.9 \\ \hline
\end{tabular}
\label{table:size_inf_time}
\vspace{-5mm}
\end{table}
\modif{\textbf{Importance of depth for the regression network.}}
\modif{In this section we study the influence of depth for the offset prediction. A regression CNN is trained using only the RGB channels as input. We compute the object pose accuracy using the ADD-S metric on 6 different objects and report the results in table \ref{table:rgbonly}. As we can see we obtain slightly better results with this method for some objects. This shows that locally an RGB patch can be more discriminant than a depth patch, this can be explained by the fact that depth patches show little details and variation compared to color. However this depends on the object. For example many patches extracted on the object “can” or “lamp” just have a slightly curved depth. The depth patches extracted from the driller show very little details compared to the RGB, contrary to the cam that shows little texture on the RGB while being textured for the depth. 
}

\begin{table}[]
\center
\caption{Comparison of ADD-S using RGB only and RGB-D.}
\begin{tabular}{|l|l|l|}
\hline
Objects & \multicolumn{1}{l|}{RGB ADD-S (\%)} & RGB-D ADD-S (\%)            \\ \hline
ape     & \textbf{91.9}                  & 91.2                      \\
cam     & 94.4                           & \textbf{95.2}             \\
can     & \textbf{99.1}                  & 98.1                      \\
drill   & \textbf{99.8}                  & 98.8 \\
lamp    & 98.5                           & \textbf{98.6}             \\
phone   & 97.6                           & \textbf{98.3}             \\ \hline
MEAN    & 96.9                           & 96.7                      \\ \hline
\end{tabular}
\vspace{-3mm}
\label{table:rgbonly}
\end{table}

\modif{\textbf{Importance of voting.}}
\modif{In this section we validate the voting approach. For two small objects we use only one $80\times 80$ patch to regress 3D points that are directly used to estimate the 3D transform. Doing so we obtain an ADD-S of 44.2\% for the ape and 51.7\% for the duck, compared to 78.2\% and 85.2\% which clearly shows the benefits of using multiple patches.}

\subsection{Visual Servoing Experiment}
\modif{To illustrate that ou approach is well suited for robotics application, }
We also propose to validate our approach within a visual servoing experiment~\cite{Chaumette06a}. \modif{Such experiment required that the pose estimation algorithm is not only precise but fast and stable over time.} We consider  a positioning task with respect to an object.
The aim of a positioning task is to reach a desired pose of the camera
{\bf r$^*$}, starting from an arbitrary initial pose. 
 We proposed to consider a position-based VS (PBVS) scheme~\cite{Chaumette06a} for which the relative pose (position and orientation) between the current and desired camera position has to be estimated. This relative pose will be estimated using the approach presented in this paper. 

The PBVS task is achieved by iteratively applying a velocity to the camera \modif{in order to minimize $\Delta \bf T$ which  is defined such that $\Delta \bf T = T^* T^{-1}$ (where both $\bf T^*$ and $\bf T$ are computed using L6DNet at the desired and current position).}
The control law is then  given by (see~\cite{Chaumette06a} for details):
\begin{equation}
 {\bf v} = - \lambda {\bf L^+} \bf \Delta r
 \label{eq:gn}
 \end{equation}
where 
$\lambda$  a positive scalar and ${\bf L^+}$ is the pseudo inverse of the interaction matrix~$\bf L$ that links the variation of the pose   to the camera velocity $\bf \tc$. 
The error is defined by ${\bf \Delta r} = ( {\bf t} ,\theta{\bf u}$), where $\bf t$ describes the translation part of the homogeneous matrix \modif{$\Delta  \bf T$} related to the transformation from the current ${\cal F}_c$ to the desired frame ${\cal F}_{c^*}$, while its rotation part ${\bf R}$ is expressed under the form $\theta{\bf u}$, where $\bf u$ represents the unit rotation-axis vector and $\theta$ the rotation angle around this axis.

\newcommand\FPuse[1]{\FPeval{\result}{#1}{\result}}
\newcommand\FPrad[1]{\FPeval{\result}{round(#1*180/pi:2)}{\result}}

Once the displacement ${\bf \Delta r}$ to be achieved is computed using our approach, it is immediate to compute the camera velocity using a classical PBVS control law~\cite{Wilson96a,Chaumette06a}:
\begin{equation}
\label{eq:pbvs}
  {\bf v} = - \lambda  \left(\begin{array}{c}  {\bf R} \; {\bf t} \\\theta{\bf u} \end{array}\right) 
\end{equation}
In such approaches, the quality of the positioning task and camera trajectory is then dependent on the quality of the estimation of the relative pose. 

\begin{figure*}
    \centering
      \includegraphics[width=0.17\textwidth]{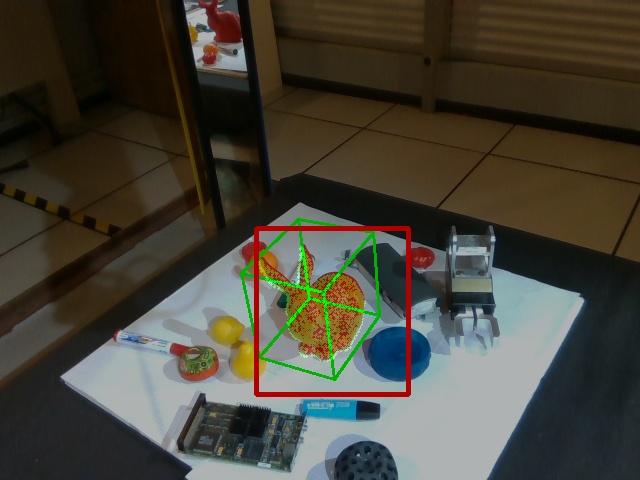}a
     \includegraphics[width=0.17\textwidth]{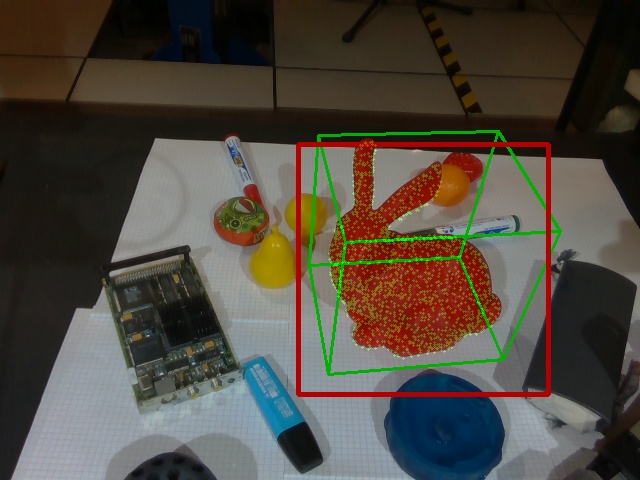}b
    \includegraphics[width=0.2\textwidth]{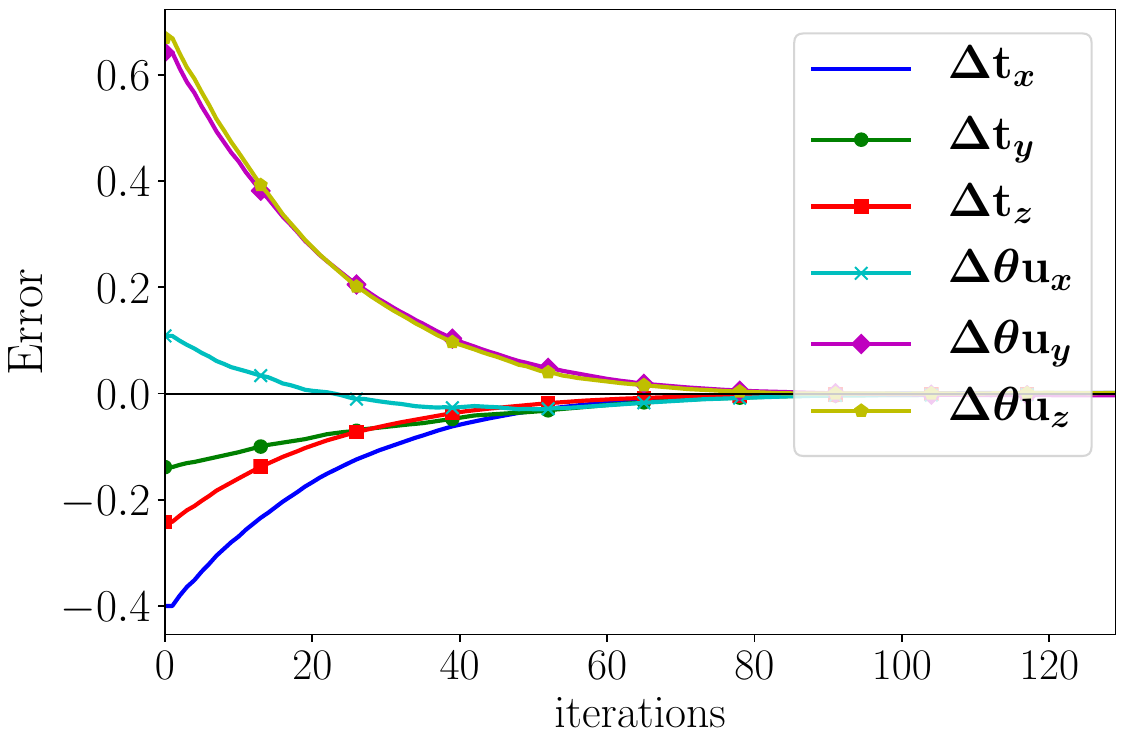}c
    \includegraphics[width=0.2\textwidth]{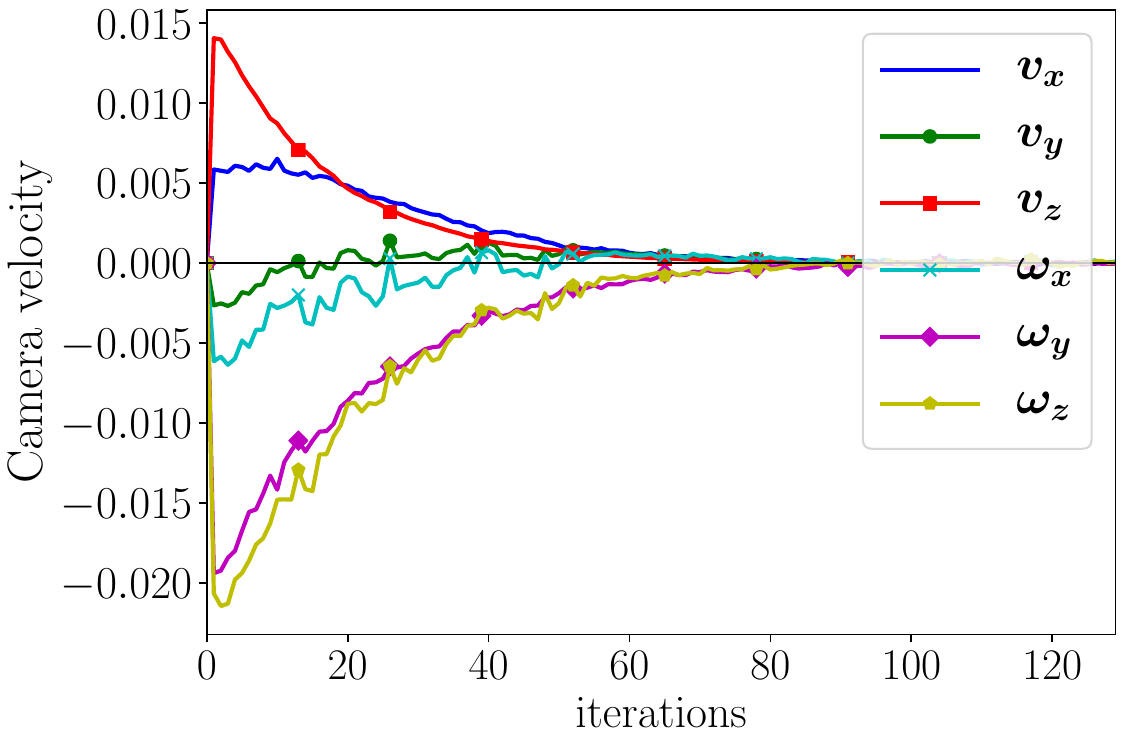}d
   \includegraphics[width=0.15\textwidth]{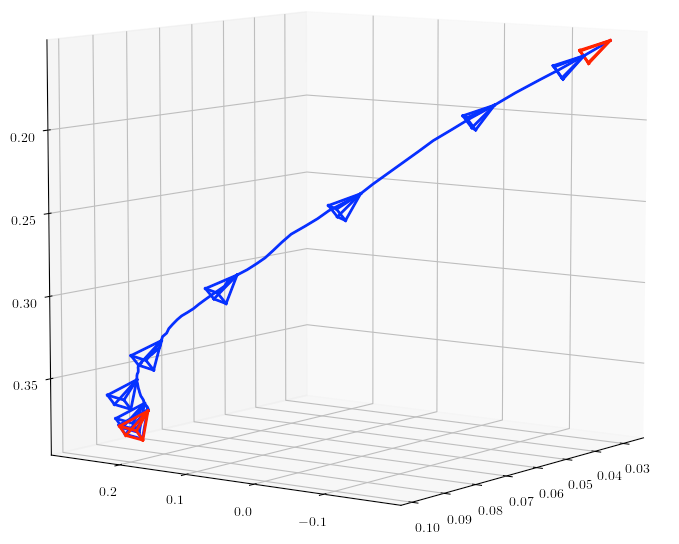}e
    \caption{Visual servoing experiment: (a) initial view (the 3D model of the target is superimposed in the image according to the estimated pose) (b) final image (c) error ${\bf \Delta r}$ in (meter and radian) (d) camera velocities (in meter/s and radian/s) (e) camera position over time.}
    \label{fig:vs}
\end{figure*}

\modif{Experiments have been carried out on a 6 DoF gantry robot, with an Intel D435 mounted on the end-effector.}
Fig.~\ref{fig:vs} illustrates the behavior of the considered VS control law. The displacement to be achieved is ${\bf \Delta r} = (-400mm, -140mm, -240mm, \FPrad{0.1085}^o, \FPrad{0.6422}^o, \FPrad{0.6695}^o) $.  The transformation between the initial and desired poses (and particularly the rotation around the $y$ and $z$ axes and translation along the $x$ and $z$ axis) is very large and makes this experiment very challenging.  The final error is ${\bf \Delta r} = (1.2mm, -1.6mm, -0.4mm, \FPrad{-0.0016}^o, \FPrad{-0.0031}^o, \FPrad{0.0012}^o)$.  \modif{Note that the evolution of the errors and of the velocities are very smooth and that the camera trajectory is  close to the expected straight line (despite a coarse eye-to-hand calibration) which demonstrated both the accuracy and efficiency of the proposed approach on long image sequences. }

\section{Conclusions}

In this paper, we introduced a novel approach to estimate 6-DoF object pose in a RGB-D image. Our method leverages the strengths of patch voting based strategies and hybrid learning-geometrical methods, using patches extracted from the image to predict a set of sparse 3D keypoints representing the object geometry in $\mathcal{F}_c$. Those points are then put in correspondence and aligned with reference keypoints to retrieve the pose. We showed that our strategy is more robust and accurate than state-of-the-art and efficient to control a camera mounted on a robot end-effector in real-time. 





\bibliographystyle{IEEEtran} 
\bibliography{IEEEabrv,IEEEexample}

\begin{thebibliography}{10}
\providecommand{\url}[1]{#1}
\csname url@rmstyle\endcsname
\providecommand{\newblock}{\relax}
\providecommand{\bibinfo}[2]{#2}
\providecommand\BIBentrySTDinterwordspacing{\spaceskip=0pt\relax}
\providecommand\BIBentryALTinterwordstretchfactor{4}
\providecommand\BIBentryALTinterwordspacing{\spaceskip=\fontdimen2\font plus
\BIBentryALTinterwordstretchfactor\fontdimen3\font minus
  \fontdimen4\font\relax}
\providecommand\BIBforeignlanguage[2]{{%
\expandafter\ifx\csname l@#1\endcsname\relax
\typeout{** WARNING: IEEEtran.bst: No hyphenation pattern has been}%
\typeout{** loaded for the language `#1'. Using the pattern for}%
\typeout{** the default language instead.}%
\else
\language=\csname l@#1\endcsname
\fi
#2}}

\bibitem{marchand2015pose}
E.~Marchand, H.~Uchiyama, and F.~Spindler, ``Pose estimation for augmented
  reality: a hands-on survey,'' \emph{IEEE Trans. on Visualization and Computer
  Graphics}, vol.~22, no.~12, pp. 2633--2651, Dec. 2016.

\bibitem{wang2019densefusion}
C.~Wang, D.~Xu, Y.~Zhu, R.~Mart{\'\i}n-Mart{\'\i}n, C.~Lu, L.~{Fei-Fei}, and
  S.~Savarese, ``Densefusion: {6D} object pose estimation by iterative dense
  fusion,'' in \emph{IEEE Conf. on Computer Vision and Pattern Recognition},
  2019, pp. 3343--3352.

\bibitem{tejani2014latent}
A.~Tejani, D.~Tang, R.~Kouskouridas, and T.-K. Kim, ``Latent-class {H}ough
  forests for {3D} object detection and pose estimation,'' in \emph{European
  Conf. on Computer Vision}, 2014, pp. 462--477.

\bibitem{hinterstoisser2012model}
S.~Hinterstoisser, V.~Lepetit, S.~Ilic, S.~Holzer, G.~Bradski, K.~Konolige, and
  N.~Navab, ``Model based training, detection and pose estimation of
  texture-less {3D} objects in heavily cluttered scenes,'' in \emph{Asian Conf.
  on Computer Vision}, 2012, pp. 548--562.

\bibitem{kehl2016deep}
W.~Kehl, F.~Milletari, F.~Tombari, S.~Ilic, and N.~Navab, ``Deep learning of
  local {RGB-D} patches for {3D} object detection and {6D} pose estimation,''
  in \emph{European Conf. on Computer Vision}, 2016, pp. 205--220.

\bibitem{rad2017bb8}
M.~Rad and V.~Lepetit, ``{BB}8: A scalable, accurate, robust to partial
  occlusion method for predicting the {3D} poses of challenging objects without
  using depth,'' in \emph{IEEE Int. Conf. on Computer Vision}, 2017, pp.
  3828--3836.

\bibitem{peng2019pvnet}
S.~Peng, Y.~Liu, Q.~Huang, X.~Zhou, and H.~Bao, ``{PVNet: Pixel-wise Voting
  Network} for {6DoF} pose estimation,'' in \emph{IEEE Conf. on Computer Vision
  and Pattern Recognition}, 2019, pp. 4561--4570.

\bibitem{tekin2018real}
B.~Tekin, S.~N. Sinha, and P.~Fua, ``Real-time seamless single shot {6D} object
  pose prediction,'' in \emph{IEEE Conf. on Computer Vision and Pattern
  Recognition}, 2018, pp. 292--301.

\bibitem{li2019cdpn}
Z.~Li, G.~Wang, and X.~Ji, ``{CDPN}: {C}oordinates-{B}ased {D}isentangled
  {P}ose {N}etwork for real-time {RGB}-{B}ased 6-{DoF} object pose
  estimation,'' in \emph{IEEE Int. Conf. on Computer Vision}, 2019, pp.
  7678--7687.

\bibitem{park2019pix2pose}
K.~Park, T.~Patten, and M.~Vincze, ``{Pix2Pose}: {P}ixel-wise coordinate
  regression of objects for {6D} pose estimation,'' in \emph{IEEE Int. Conf. on
  Computer Vision}, 2019, pp. 7668--7677.

\bibitem{li2018deepim}
Y.~Li, G.~Wang, X.~Ji, Y.~Xiang, and D.~Fox, ``{DeepIM}: {Deep Iterative
  Matching} for {6D} pose estimation,'' in \emph{European Conf. on Computer
  Vision}, 2018, pp. 683--698.

\bibitem{xiang2017posecnn}
Y.~Xiang, T.~Schmidt, V.~Narayanan, and D.~Fox, ``{PoseCNN}: A convolutional
  neural network for {6D} object pose estimation in cluttered scenes,'' in
  \emph{{Robotics: Science and Systems (RSS)}}, 2018.

\bibitem{tekin2019h+}
B.~Tekin, F.~Bogo, and M.~Pollefeys, ``{H+O}: Unified egocentric recognition of
  {3D} hand-object poses and interactions,'' in \emph{IEEE Conf. on Computer
  Vision and Pattern Recognition}, 2019, pp. 4511--4520.

\bibitem{he2020pvn3d}
Y.~He, W.~Sun, H.~Huang, J.~Liu, H.~Fan, and J.~Sun, ``Pvn3d: A deep point-wise
  3d keypoints voting network for 6dof pose estimation,'' in \emph{IEEE Conf.
  on Computer Vision and Pattern Recognition}, 2020, pp. 11\,632--11\,641.

\bibitem{riegler2013hough}
G.~Riegler, D.~Ferstl, M.~R{\"u}ther, and H.~Bischof, ``Hough networks for head
  pose estimation and facial feature localization,'' \emph{Journal of Computer
  Vision}, vol. 101, no.~3, pp. 437--458, 2013.

\bibitem{fanelli2011real}
G.~Fanelli, J.~Gall, and L.~Van~Gool, ``Real time head pose estimation with
  random regression forests,'' in \emph{IEEE Int. Conf. on Computer Vision and
  Pattern Recognition}, 2011, pp. 617--624.

\bibitem{kacete2016real}
A.~Kacete, J.~Royan, R.~Seguier, M.~Collobert, and C.~Soladie, ``Real-time eye
  pupil localization using {H}ough regression forest,'' in \emph{IEEE Winter
  Conf. on Applications of Computer Vision}, 2016.

\bibitem{gall2011hough}
J.~Gall, A.~Yao, N.~Razavi, L.~Van~Gool, and V.~Lempitsky, ``Hough forests for
  object detection, tracking, and action recognition,'' \emph{IEEE Trans. on
  PAMI}, vol.~33, no.~11, pp. 2188--2202, 2011.

\bibitem{do2018deep}
T.-T. Do, M.~Cai, T.~Pham, and I.~Reid, ``{Deep-6DPose}: Recovering {6D} object
  pose from a single {RGB} image,'' \emph{arXiv preprint arXiv:1802.10367},
  2018.

\bibitem{kehl2017ssd}
W.~Kehl, F.~Manhardt, F.~Tombari, S.~Ilic, and N.~Navab, ``{SSD-6D}: Making
  {RGB}-based {3D} detection and {6D} pose estimation great again,'' in
  \emph{IEEE Int. Conf. on Computer Vision}, 2017, pp. 1521--1529.

\bibitem{sundermeyer2018implicit}
M.~Sundermeyer, Z.-C. Marton, M.~Durner, M.~Brucker, and R.~Triebel, ``Implicit
  {3D} orientation learning for {6D} object detection from {RGB} images,'' in
  \emph{European Conf. on Computer Vision}, 2018, pp. 699--715.

\bibitem{mahendran20173d}
S.~Mahendran, H.~Ali, and R.~Vidal, ``{3D} pose regression using convolutional
  neural networks,'' in \emph{IEEE Int. Conf. on Computer Vision}, 2017, pp.
  2174--2182.

\bibitem{pavlakos20176}
G.~Pavlakos, X.~Zhou, A.~Chan, K.~G. Derpanis, and K.~Daniilidis, ``{6-DoF}
  object pose from semantic keypoints,'' in \emph{IEEE Int. Conf. on Robotics
  and Automation}, 2017, pp. 2011--2018.

\bibitem{wu2016single}
J.~Wu, T.~Xue, J.~J. Lim, Y.~Tian, J.~B. Tenenbaum, A.~Torralba, and W.~T.
  Freeman, ``Single image 3d interpreter network,'' in \emph{European Conf. on
  Computer Vision}.\hskip 1em plus 0.5em minus 0.4em\relax Springer, 2016, pp.
  365--382.

\bibitem{grabner20183d}
A.~Grabner, P.~M. Roth, and V.~Lepetit, ``3d pose estimation and 3d model
  retrieval for objects in the wild,'' in \emph{IEEE Conf. on Computer Vision
  and Pattern Recognition}, 2018, pp. 3022--3031.

\bibitem{comaniciu2002mean}
D.~Comaniciu and P.~Meer, ``Mean shift: A robust approach toward feature space
  analysis,'' \emph{IEEE Trans. on PAMI}, vol.~24, no.~5, pp. 603--619, 2002.

\bibitem{cheng1995mean}
Y.~Cheng, ``Mean shift, mode seeking, and clustering,'' \emph{IEEE Trans. on
  PAMI}, vol.~17, no.~8, pp. 790--799, 1995.

\bibitem{arun1987least}
K.~S. Arun, T.~S. Huang, and S.~D. Blostein, ``Least-squares fitting of two
  {3-D} point sets,'' \emph{IEEE Trans. on PAMI}, no.~5, pp. 698--700, 1987.

\bibitem{fitzgibbon2003robust}
A.~W. Fitzgibbon, ``Robust registration of {2D} and {3D} point sets,''
  \emph{Image and vision computing}, vol.~21, no. 13-14, pp. 1145--1153, 2003.

\bibitem{malis2006experiments}
E.~Malis and E.~Marchand, ``Experiments with robust estimation techniques in
  real-time robot vision,'' in \emph{IEEE/RSJ Int. Conf. on Intelligent Robots
  and Systems}, 2006, pp. 223--228.

\bibitem{besl1992method}
P.~J. Besl and N.~D. McKay, ``Method for registration of 3-{D} shapes,'' in
  \emph{Sensor fusion IV: control paradigms and data structures}, vol. 1611,
  1992, pp. 586--606.

\bibitem{tensorflow2015-whitepaper}
\BIBentryALTinterwordspacing
M.~Abadi, \emph{et~al.}, ``{TensorFlow}: Large-scale machine learning on
  heterogeneous systems,'' 2015. [Online]. Available:
  \url{http://tensorflow.org/}
\BIBentrySTDinterwordspacing

\bibitem{redmon2018yolov3}
J.~Redmon and A.~Farhadi, ``Yolov3: An incremental improvement,'' \emph{arXiv
  preprint arXiv:1804.02767}, 2018.

\bibitem{Chaumette06a}
F.~Chaumette and S.~Hutchinson, ``Visual servo control, {P}art~{I}: Basic
  approaches,'' \emph{IEEE Robotics and Automation Magazine}, vol.~13, no.~4,
  pp. 82--90, December 2006.

\bibitem{Wilson96a}
W.~Wilson, C.~Hulls, and G.~Bell, ``Relative end-effector control using
  cartesian position-based visual servoing,'' \emph{IEEE Trans. on Robotics and
  Automation}, vol.~12, no.~5, pp. 684--696, Oct. 1996.

\end{thebibliography}

\newpage
\section*{Appendix}
We report here the histograms of regression error for all objects and the impact of patch size on both true positive rate of classification and regression error.
~\\
\begin{table}[h]
\center
\caption{True positive rate of classification (\%) for different patch sizes and objects.}
\begin{tabular}{|l|llll|}
\hline
Patch size & 32   & 48   & 64   & 80   \\ \hline
ape        & 98.5 & 96,9 & 98,2 & 98.3 \\ 
cam        & 90.1 & 85   & 92,8 & 96.6 \\ 
duck       & 97.8 & 92,5 & 91,3 & 97.0 \\ 
drill      & 94.5 & 94   & 95,1 & 91.2 \\ \hline
\end{tabular}
\end{table}

\begin{table}[h]
\centering
\caption{Regression error (mm) for different patch sizes and objects.}
\begin{tabular}{|l|llll|}
\hline
Patch size & 32   & 48   & 64   & 80   \\ \hline
ape        & 9    & 10,7 & 7,2  & 11.3 \\ 
cam        & 33,6 & 27,9 & 16,9 & 17,8 \\ 
duck       & 11,5 & 9,1  & 10,1 & 9,8  \\ 
drill      & 15,5 & 12,6 & 12,8 & 16,5 \\ \hline
\end{tabular}
\end{table}

\begin{figure}
    \centering
      \includegraphics[width=0.92\columnwidth]{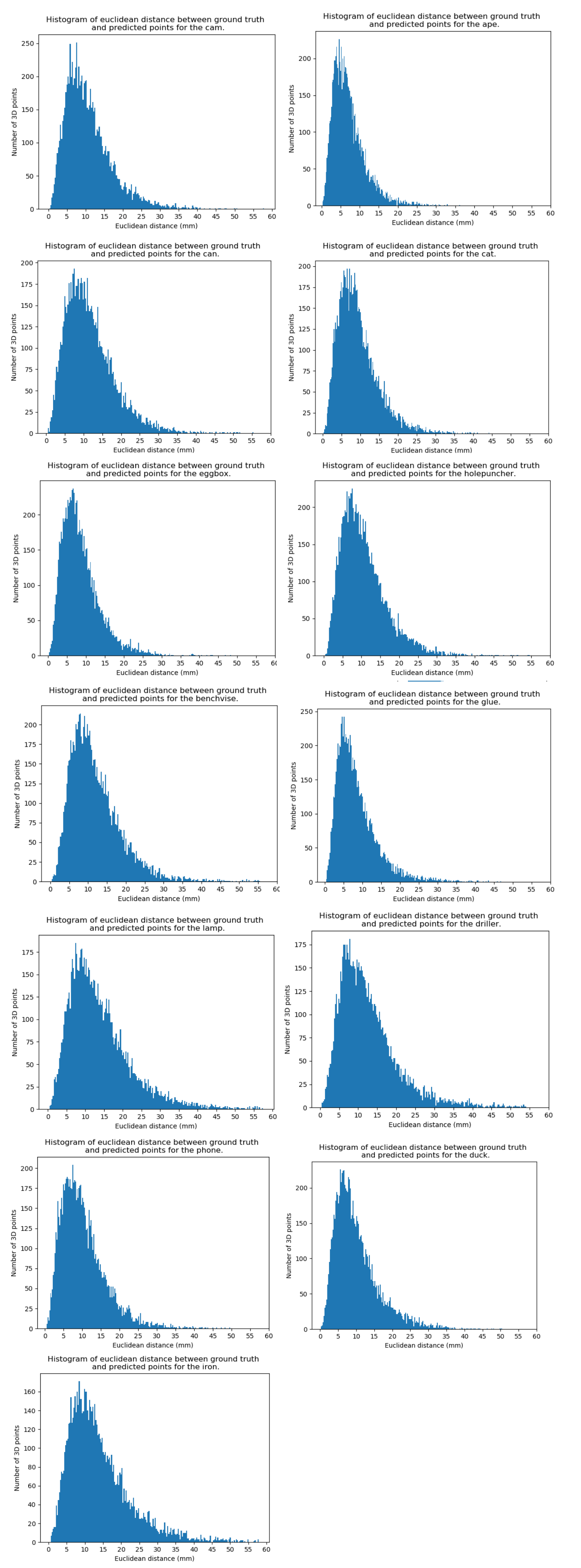}
    \caption{Histograms of regression error for each LineMod object.}
    \label{hist}
\end{figure}

\end{document}